\documentclass[10pt,twocolumn,letterpaper]{article}

\usepackage{cvpr}
\usepackage{times}
\usepackage{epsfig}
\usepackage{graphicx}
\usepackage{amsmath}
\usepackage{amssymb}

\usepackage{booktabs}       
\usepackage{algorithm}
\usepackage{algpseudocode}
\usepackage[square,sort,comma,numbers]{natbib}
\usepackage{color}
\usepackage{pifont} 

\usepackage{caption} 
\usepackage{nicefrac}       
\usepackage{import}
\usepackage{color}
\usepackage{transparent}
\graphicspath{ {figures/} }

\newcommand{\xmark}{\ding{55}}%
\DeclareMathOperator*{\argmin}{arg\,min}

\usepackage[pagebackref=true,breaklinks=true,colorlinks,bookmarks=false]{hyperref}

\cvprfinalcopy 

\ifcvprfinal\fi

\begin{document}

\title{Transition Forests: Learning Discriminative Temporal Transitions for Action Recognition and Detection}

\author{Guillermo Garcia-Hernando\\
Imperial College London\\
{\tt\small ggarciah@imperial.ac.uk}
\and
Tae-Kyun Kim\\
Imperial College London\\
{\tt\small tk.kim@imperial.ac.uk}
}

\maketitle
\thispagestyle{empty}

\begin{abstract}
  A human action can be seen as transitions between one's body poses over time, where the transition depicts a temporal relation between two poses. Recognizing actions thus involves learning a classifier sensitive to these pose transitions as well as to static poses. In this paper, we introduce a novel method called \textit{transitions forests}, an ensemble of decision trees that both learn to discriminate static poses and transitions between pairs of two independent frames. During training, node splitting is driven by alternating two criteria: the standard classification objective that maximizes the discrimination power in individual frames, and the proposed one in pairwise frame transitions. Growing the trees tends to group frames that have similar associated transitions and share same action label incorporating temporal information that was not available otherwise. Unlike conventional decision trees where the best split in a node is determined independently of other nodes, the transition forests try to find the best split of nodes jointly (within a layer) for incorporating distant node transitions. 
   When inferring the class label of a new frame, it is passed down the trees and the prediction is made based on previous frame predictions and the current one in an efficient and online manner.
   We apply our method on varied skeleton action recognition and online detection datasets showing its suitability over several baselines and state-of-the-art approaches.
\end{abstract}

\section{Introduction}
\label{introduction}
Recognizing and localizing human actions is an important and classic problem in computer vision \cite{aggarwal2011human,de2016online} with a wide range of applications including pervasive health-care, robotics, game control, etc. With recently introduced cost-effective depth sensors and reliable real-time body pose estimation \citep{shotton2013real}, skeleton-based action recognition has become popular because of the advantage of pose features over raw RGB video approaches in both accuracy and efficiency \cite{yao2011does}.

Popular approaches for action recognition and localization include using generative models such as state-space models  \citep{wu2014leveraging,lehrmann2014efficient}; or tackling it as a classification problem of either the whole sequence \citep{vemulapalli2014human,zhu2013fusing}, a small chunk of frames \citep{Zanfir_2013_ICCV,msrc12} or deep recurrent models \cite{du2015hierarchical, li2016online}. The best performing methods focus either on modelling the temporal dynamics using time-series models \citep{zhangefficient} or recognizing key-poses \citep{zhu2016human}, showing that both static and dynamic information are important cues for actions. Motivated by this, we consider decision forests \cite{breiman2001random}, which have been widely adopted in computer vision \cite{shotton2013real,yao2011does,tang2013real}, owing to many desired properties: clusters obtained in leaf nodes, scalability, robustness to overfitting, multiclass learning and efficiency.

The main challenge of using decision forests for temporal problems lies in dealing with temporal dependencies. Previous approaches encode the temporal variable in the feature space by stacking multiple frames \cite{msrc12}, handcrafting temporal features \citep{yu2010real,zhu2013fusing} or creating codebooks \cite{yu2010real}.   However, these methods require the temporal cues to be explicitly given instead of automatically learning them. Attempting to relieve this,  \citep{yao2011does,WACV} add a temporal regression term and frames individually vote for an action center, breaking the temporal continuity and thus not fully capturing the temporal dynamics. \citep{lehrmann2014efficient} proposed a generative state-space without exploiting the benefit of having rich labelled data. \citep{dapogny2015pairwise} groups pairs of distant frames and grows trees using handcrafted split functions to cover different label transitions, with the difficulty of designing domain-specific functions and making the model complexity to increase with the number of labels. 

In this work, we propose `transition forests', an ensemble of randomized tree classifiers that learns both static pose information and temporal transitions in a discriminative way. Temporal dynamics are learned while training the forest (besides any temporal dependencies in the feature space) and predictions are made by taking into account previous predictions. Introducing previous predictions makes the learning problem more challenging as a consequence of the ``chicken and egg" problem: making a decision in a node that depends on the decision in other nodes and vice versa. To tackle this problem, we propose a training procedure that iteratively groups pairs of frames that have similar associated frame transitions and class label in a given level of the tree. We combine both static and transition information by randomly assigning nodes to be optimized by classification or transition criteria. In the end of tree growth, training frames arriving at leaf nodes represent effectively a class label and associated transitions. We found that adding such temporal relation in training helped to obtain more robust single frame predictions. Using single frames helped us in keeping the complexity low and being able to make online predictions, two crucial conditions to make our approach applicable to real life scenarios. 

\section{Related work}
\label{Sec:related work}

\textbf{Skeleton-based action recognition.} Generative models \citep{xia2012view,wu2014leveraging,lehrmann2014efficient} such as Hidden Markov Models (HMM) have been proposed with the disadvantages of being difficult to estimate model parameters and time consuming learning and inference stages. Discriminative approaches have been widely adopted due to their superior performance and efficiency. 
For instance, \citep{action3d} extracts local features from body joints captures temporal dynamics using Fourier Temporal Pyramids (FTP), further classifying the sequence using Support Vector Machines (SVM).
Similarly, \citep{vemulapalli2014human,rolling} represents the whole skeletons as points in a Lie group before temporally aligning sequences using Dynamic Time Warping (DTW) and applying FTP.
 \citep{Zanfir_2013_ICCV} proposes a Moving Pose descriptor (MP) using both pose and atomic motion information and then temporally mining key frames using a k-NN aproach in contrast to \citep{jung2014enhanced} that uses DTW. 
Using key frames or key motion units has been also studied by \cite{devanne2015combined,wang2016mining,zhu2016human} showing good performance revealing that static information is important to recognize actions. Recently, deep models using Recurrent Neural Networks (RNN) \citep{du2015hierarchical} and Long-Short Term Memory (LSTM) \cite{Veeriah_2015_ICCV,zhu2016co} have been proposed to model temporal dependencies, but showed inferior performance than recent (offline) models that explicitly exploit static information \cite{wang2016mining,wang2016graph} or well-suited time-series mining \citep{zhangefficient}. Our forest learns bost static per-frame and temporal information in a discriminative way. 

\textbf{Skeleton-based online action detection.} Detecting actions on streaming data \cite{de2016online} has been less explored than recognizing segmented sequences, while being more interesting in real scenarios. Early approaches \citep{msrc12} include using short sequences of frames or short motion information \citep{Zanfir_2013_ICCV} to vote if an action is being performed. A similar approach but adding multi-scale information was proposed by \cite{sharaf2015real}, while \cite{meshry2016linear} proposed a dynamic bag of features. Recently, \cite{li2016online} introduced a more realistic dataset, baseline methods and shown state-of-the-art performance with a classification/regression RNN, later improved by \cite{baek2016real} with the use of RGB-D spatio-temporal contexts and decision forests.

\textbf{Forests and temporal data.} Standard forest approaches for action recognition such as \citep{msrc12} directly stack frames and grows forests to classify them. \cite{zhu2013fusing,florence3d} create bags of poses and classified the whole sequences. Using the clustering properties of trees, \citep{yu2010real} construct codebooks with the help of different heuristic rules capturing structural information. These approaches require the temporal cues to be directly encoded in the feature space. To relieve this, \citep{yao2011does,yu2013unconstrained,cviu16} add a temporal regression term and maps appearance and pose features to vote in an action Hough space. \citep{WACV} proposes Trajectory Hough Forest (THF) that computes histograms of tree paths over consecutive color and flow trajectory patches and uses them as weights for prediction. However, in Hough frameworks, temporal information is captured as temporal offsets with respect to a temporal center of independent samples, breaking the temporal continuity and requiring the whole sequence to be observed. On the contrary, we explicitly capture the rich temporal dynamics and are able to perform online predictions. \citep{dapogny2015pairwise} proposes Pairwise Conditional Random Forests (PCRF) for facial expression recognition consisting of trees of which handcrafted split functions operate on pairs of frames. These pairs are formed to cover different facial dynamics and fed into multiple subsets of decision trees that are conditionally drawn based on different label transitions, making the ensemble size proportional to the number of labels.  
By contrast, our layer-wise optimization tries to automatically learn the best node splits based on single frames maximizing both static and transition information within the same tree and thus not needing handcrafted split functions or to create different trees based on different labels. Generative methods based on forests include Dynamic Forest Models (DFM) \citep{lehrmann2014efficient}, which are ensembles of autoregressive trees that store multivariate distributions at their leaf nodes. These distributions model observation probabilities given short history of previous $k$ frames. Similar to HMM, a decision forest is trained for each action label and inference is performed maximizing likelihood of the observed sequence. Recently, \cite{chen2016learning} proposed to learn smooth temporal regressors for real time camera planning. We share with \cite{chen2016learning} the recurrent nature of making online predictions conditioned on our own previous predictions, however our approach differs in how the recurrency is defined in both learning and inference stages. We compare some relevant methods in Section \ref{Sec: Experiments}.

\textbf{Tree-based methods for structured prediction.} A related line of work \citep{shotton2008semantic,nowozin2011decision,kontschieder2013geof,shotton2013decision} proposes decision forests methods for image segmentation. The objective of these approaches is to obtain coherent pixel labels and, in order to connect multiple pixel predictions, decision forests are linked with probabilistic graphical models. While these methods focus on the  spatial coherence of predictions in an image space, our method tries to capture discriminative changes of data/prediction in a temporal domain. 

\section{Transition forests}
\label{Sec:trans}

Suppose we are given a training set $S$ composed of temporal sequences of input-output pairs $\{(x_1,y_1),...,(x_t,y_t)\}$ where $x_t$ is a frame feature vector encoding pose information and $y_t$ is its corresponding action label (or background in detection setting). Our objective is to infer $y_t$ for every given $x_t$ using decision trees.  On a decision tree, an input instance $x_t$ starts at the root and traverses different internal nodes until it reaches a leaf node.  Each internal node $i \in \mathcal{N}$  contains a binary split function $f$ with parameters $\theta_i$ deciding whether the instance should be directed to the left or to the right child nodes.

\begin{figure}[!t]
\def\svgwidth{\columnwidth}
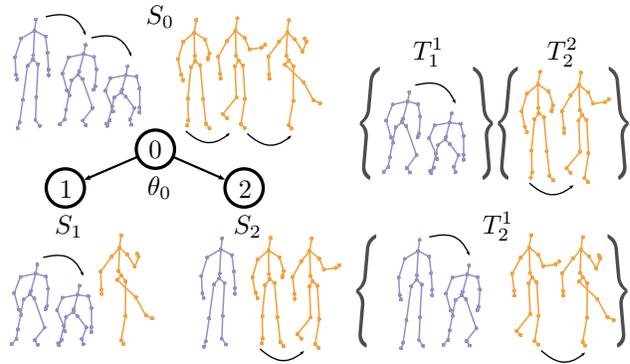
\caption{Consecutive frames representing two different actions (in purple `duck', in orange `kick') arrive at node $0$. These frames are split in two different subsets $S_1$ and $S_2$ corresponding to child nodes $1$ and $2$. We compute the transitions as pairs of $d$-distant frames ($d=1$ in this example) and we group them according to the route of each individual frame. $T_1^1$ and $T^2_2$ present only one transition, while $T_2^1$ two (one per class) and  $T_1^2$ is empty. $T_i^j$ are determined by $\theta_0$.}
\label{fig:trans_figure}	
\end{figure}

Consider the set of nodes $\mathcal{N}_l \subset \mathcal{N}$ at a level $l$ of a decision tree. Let $S_i$ denote the set of labeled training instances $(x_t,y_t)$ that reached node $i$ (see Fig. \ref{fig:trans_figure}). For each pair of nodes $i,j \in \mathcal{N}_l$, we can compute the set of pairs of frames $T_{i}^{j}$ that travel from node $i$ to node $j$ in $d$ time steps as: 
\begin{multline}\label{transprob}
T_{i}^{j}=\{\{(x_{t-d},y_{t-d}),(x_t,y_t)\}  \mid \\  (x_{t-d},y_{t-d})  \in S_{i} \wedge  (x_t,y_t) \in S_j\}\,,
\end{multline}
where we term the set of pairs of frames $T_{i}^{j}$ as \textit{transitions from node i to j}. Note that  $T_{i}^{j}$ depends on frames that reached nodes $i$ and $j$ and time distance $d$.  In order to capture different temporal patterns, we vary the distance $d$ from one to a $k$-distant frame. In the following, we will refer to parameter $k$ as the \textit{temporal order of the transition forest}.

In the example shown in Fig. \ref{fig:trans_figure} we observe that the decision $f(\theta_0,S_0)$ is quite good as it separates $S_0$ in two sets, $S_1$ and $S_2$, in which one action label predominates. If we examine the transitions associated to this split, we see that we obtain two pure sets, $T_1^1$ and $T^2_2$, one mixed set $T_2^1$ and one empty set $T_1^2$. Imagine now that we observe the `kick' frame in $S_1$ and we would have to make a decision based on this split, we would certainly assign the wrong label `duck' with an uncertainty of $2/3$. Alternatively, if we check the previous observed frame (in $S_2$) and inspect its associated transition $T_2^1$, the uncertainty is now $1/2$ and thus we would be less inclined to make a wrong decision. 

From the above example, we deduce that if we had obtained a better split and both child nodes were pure, we would certainly make a good decision by only looking at child nodes. However, good splits are difficult to learn if the temporal dynamics are not well captured on the feature space. On the other hand, if we had obtained a split that made transitions pure, we could also make a good decision. These observations motivate us to study how learning transitions between frames can help us to improve our predictions by introducing temporal information that was not available otherwise.
\subsection{Learning transition forests} \label{Subsec: learning}
Our method for training a transition tree works by growing a tree one level at a time similar to \citep{shotton2013decision}. At each level, we randomly assign one splitting criterion to each node, choosing between classification and transition. The classification criterion maximizes the class separation of static poses while the transition criterion groups frames that share similar transitions.
As mentioned above, in order to maximize the span of temporal information learned, we learn transitions between $d$-distant pairs of frames  (Eq. \ref{transprob}) from  previous frame up to the temporal order of the forest,  $k$. For each tree, we randomly assign a value of $d$ in the mentioned range and we keep it constant during the growth of this particular tree. For a total ensemble of $\mathcal{M}$ trees we will have subsets of trees trained with different $d$ value: $\mathcal{M} = \mathcal{M}_1\cup...\cup \mathcal{M}_k$.

Consider a node $i \in \mathcal{N}_l$ and a decision $\theta_i$. According to $\theta_i$, the instances in $S_i$ are directed to its left or right child nodes, $2i+1$ and $2i+2$ respectively, as $S_{2i+1}=\{(x_t,y_t) \in S_i \mid f(\theta_i,x_t) \leq 0\}$ and $S_{2i+2}=S_ i\setminus S_{2i+1}$. Note that the split function $f$ operates on a single frame, which will be shown important in the inference stage. After splitting, we can compute the sets of transitions between their child nodes  $\{2i+1, 2i+2\} \subseteq \mathcal{N}_{l+1}$ as  $\{T_{2i+m}^{2i+n}\}_{m,n\in\{1,2\}}$. Note that $T_i^i$ is split in four disjoints sets, each one related to the combination of transitions associated to its child nodes. The decision $\theta_i$ is chosen based on the minimization of an objective function.

\begin{figure*}[!t]
\begin{center}
\def\svgwidth{\textwidth}
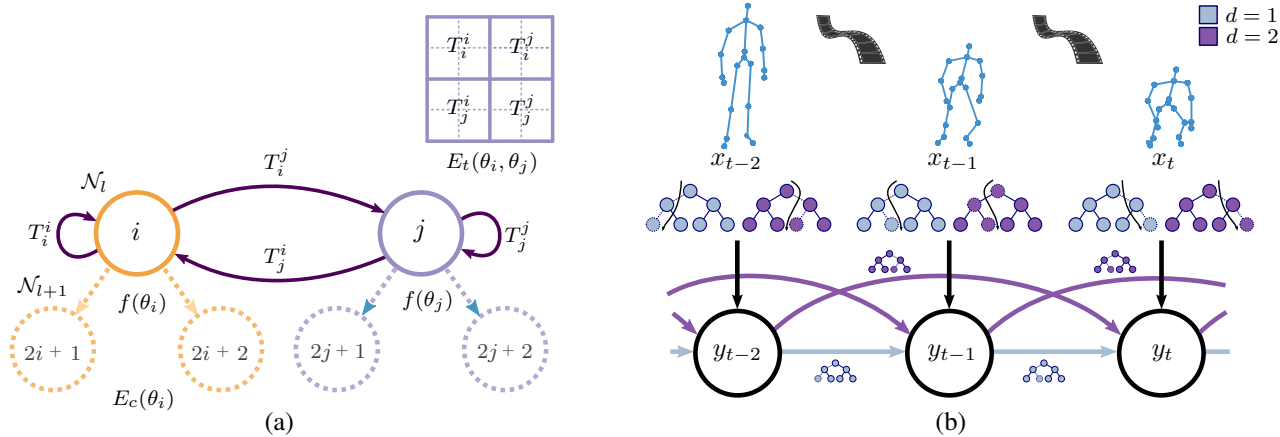
\end{center}
   \caption{(a) Growing a level $l$ of a transition tree depends on all the node decisions $\theta_i$ and $\theta_j$ at the same time. Each $T_i^j$ divides in four disjoint sets according to the different routes that a pair of samples can follow. (b) In inference, each individual frame is passed down the forest and static pose classification is combined with transition probability. Transition probability is computed using the trees trained for specific $d$-distant frames (shown in different color). In this example $k=2$ and $|\mathcal{M}|=2$.  }
\label{fig:training}	
\end{figure*}

\textbf{Objective function.} The objective function has two associated terms: one for single frame classification $E_c$ and one for transitions between child nodes denoted as $E_t$. The classification term $E_c$ is the weighted Shannon entropy of the class distributions over the set of samples that reach the child nodes $\{S_{2i+m}\}_{m \in \{1,2\}}$ as in standard classification forests. Willing to decrease the uncertainty of transitions while growing the tree, the transition term aims to learn node decisions in a way that subsets of transitions are more pure in the next level. For a node $j \in \mathcal{N}_l$, the transition term is a function of the transitions between its child nodes and it is defined as:
\begin{equation}
E_t(\theta_j) = \sum_{ m,n\in\{1,2\}} |T_{2j+m}^{2j+n}|H(T_{2j+m}^{2j+n})\,,
\label{eq:et}
\end{equation}
where $T_{(\cdot)}^{(\cdot)}$ is defined in Eq. (\ref{transprob}) and $H(T_{(\cdot)}^{(\cdot)})$ is the Shannon entropy computed over the different label transitions. These two terms could be alternated or weighted-summed as single node optimizations. However, in order to reflect transitions between more distant nodes and capture further temporal information, we extend $E_t$ to consider the set of all available nodes in a given level of a tree (as shown in Fig. \ref{fig:training} (a)). For this, we randomly assign a subset of parent nodes $N_c$ and $N_t$ to be optimized by $E_c$ and $E_t$ respectively. Given that transitions between nodes depend on the split decisions at different nodes, the task of learning a level can be formulated as the joint minimization of an objective function over the split parameters associated to the level nodes as: 
\begin{equation} \label{eq:objective}
 \min_{\{\theta_i\}}  E_c(\{\theta_{i}\}_{i \in N_c})+E_t(\{\theta_i\}_{i \in N_c \cup N_t })\,.
\end{equation}

\textbf{Optimization.} The problem of minimizing the objective function (Eq. \ref{eq:objective}) is hard to solve. One could think of randomly assign values to  $\{\theta_i\}$ and pick the values that minimize the objective in a similar way to standard greedy optimization in decision trees. However, the search space grows exponentially with the depth of the tree and evaluating $E_t$ for all nodes and samples at the same time is computationally expensive. Our strategy to relieve these problems is presented in Algorithm \ref{alg:optimization}.  Given that $E_c$ only depends on decisions in $N_c$ nodes, we can optimize these nodes using the standard greedy procedure. Once optimized and fixed all nodes in $N_c$, we iterate over every node in $N_t$ to find the split function that minimizes a local version of $E_t$, denoted as $E'_t$, that keeps all the split parameters fixed except the one of the considered node. It is defined for a node $j \in N_t$ and it depends on the transitions between its child nodes and all the transitions \textit{from} and \textit{to} these child nodes:
\begin{multline}\label{transition_ob}
  E'_t(\theta_j|\{\theta_i\}_{i\neq j \in N_c \cup N_t})= \sum_{m,n\in\{1,2\}}\overbrace{ |T_{2j+m}^{2j+n}|H(T_{2j+m}^{2j+n})}^{\text{between j's child nodes (c.n.)}} \\+ \sum_{\substack{ i \\ m,n\in\{1,2\}}} \underbrace{|T_{2j+m}^{2i+n}|H(T_{2j+m}^{2i+n})}_{\text{from j's c.n. to i's c.n.}}+\underbrace{|T_{2i+m}^{2j+n}|H(T_{2i+m}^{2j+n})}_{\text{to j's c.n. from i's c.n.}}\,.
\end{multline}
The value of $E'_t$ decreases (or does not change) at each iteration, thus indirectly minimizing $E_t$. Following this strategy it is not likely to reach a global minimum, but in practice we found that is effective to our problem. Note that computing Eq. \ref{transition_ob} needs the split parameters in other nodes to be available, forcing us to initialize them before the first iteration. We found that an initialization of nodes using $E_c$ helped the algorithm to converge faster than using a random initialization relieving us of computational cost.

\begin{algorithm}[!t]
	\caption{Learning level $l$ of a transition tree}
	\label{alg:optimization}

	\begin{algorithmic}[1]
	\Require Set of nodes $\mathcal{N}_l$ at level $l$ and temporal order $d$
	\Ensure Set of split function parameters $\{\theta_{i}\}$
	\Procedure{LearnLevel}{$\mathcal{N}_l$}
		\State randomly assign nodes in $\mathcal{N}_l$ to $N_c$ and $N_t$
		\ForAll{$i \in N_c$}				
			\State optimize $N_c$ using $E_c$
			\State save and fix $\theta_{i}$
	    \EndFor

		\State initialize $\{\theta_{j}\}$ for $j \in N_t$
		\While{something changes}
		\ForAll{$j \in N_t$}
			\State $\Theta \gets $ random feature/threshold selection
			\State $\theta_{j} \gets \argmin_{\theta' \in \Theta} E'_t(\theta_j|\{\theta_i\}_{i\neq j \in N_c \cup N_t}  )$			
		\EndFor
		\EndWhile
	\EndProcedure
\end{algorithmic}
\end{algorithm}

\subsection{Inference}

Restricting ourselves to the set of leaf nodes $\mathcal{L}$, we assign each transition subset $\{T_{i}^{j}\}_{i,j\in\mathcal{L}}$ a conditional probability distribution over label transitions denoted $\pi_i^j(y_t|y_{t-d})$. %
This is different from classification forests where the \textit{classification probability} $\pi_i(y_t)$ is estimated over all the set of training instances $S_i$ that reached the leaf node $i$. Instead, we focus on subsets of transitions that depend on the leaf node (prediction) that previous $d$-distant frame reached. Note that the split function $f$ is defined for a single frame, enabling us to perform individual frame predictions.
For an ensemble of $\mathcal{M}_d$ transition trees, we define a prediction function given two $d$-distant frames:
\begin{equation}\label{eq:trans}
p_d(y_t|x_t,x_{t-d},y_{t-d}) = \frac{1}{|\mathcal{M}_d|} \sum_{ m \in \mathcal{M}_d } (\pi_{\ell(x_t)}^{\ell(x_{t-d})}(y_t|y_{t-d}))^{(m)}\,,
\end{equation}
where $\ell(x_t)$ and $\ell(x_{t-d})$ are the leaf nodes reached by $x_t$ and $x_{t-d}$ at $m$-th tree respectively. We name this probability as \textit{transition probability}. We combine the transition probability for different previous pairs of frames up to $k$ with the \textit{classification probability} (see Fig. \ref{fig:training} (b)). Combining the static classification probability with the temporal transition probability defines our final prediction equation for a transition forest of temporal order $k$: 
\begin{multline}
\label{eq:prediction}
p(y_t|x_t,x_{t-1},...,x_{t-k},y_{t-1},...,y_{t-k}) = \\ \frac{1}{|\mathcal{M}|} \sum_{ m} (\pi_{\ell(x_t)}(y_t))^{(m)}\frac{1}{k}\sum_{1 \leq d \leq k} p_d(y_t|x_t,x_{t-d},y_{t-d})\,.
\end{multline}

For each frame $x_t$ we obtain a probability  of the frame belonging to one action (plus background in detection setting) based on $k$ previous predictions. In the \textit{action recognition} setting we average the per-frame results to predict the whole sequence. On the other hand, for \textit{online action detection}, we define two thresholds, $\beta_s$ and $\beta_e$, to locate the start and the end frame of the action. When the score for one action exceeds $\beta_s$, we aggregate the results since the start of the action and we do not allow any action change until the score is less than $\beta_e$. 

\subsection{Implementation details}

If the training data is not enough, we may encounter empty transition subsets at low levels of the tree. For this reason, we set a minimum number of instances needed to estimate their probability distribution and we empirically set this parameter to ten in our experiments.  This parameter is conceptually the same as the stopping criterion of requiring a minimum number of samples to keep splitting a node. 


\section{Experimental evaluation}\label{Sec: Experiments}

In the following we present experiments to evaluate the effectiveness of our approach. We start evaluating our approach for action recognition and we follow with online action detection. In all experiments we performed standard pre-processing on given joint positions similar to \citep{vemulapalli2014human} making them invariant to scale, rotation and point of view. 

\subsection{Baselines}
We compare our approach with five different forest-based baselines detailed next. For fair comparison, we always use the same number of trees in all methods and we adjust the maximum depth for best performance.

\textbf{Random Forest \cite{breiman2001random} (RF).} To assess how well performs a decision forest while only using static information, we implement a single frame-based random forest only using $E_c$.

\textbf{Sliding Window Forest \citep{msrc12} (SW).} To compare our learning of temporal dynamics with the strategy of stacking multiple frames, we implement a forest using the sliding window setting in which the temporal order $k$ the number of previous frames in the window.  

\textbf{Trajectory Hough Forest \citep{WACV} (THF).} To compare with a temporal regression method, we implement \citep{WACV} and adapt their color trajectories to poses and their histograms to deal with a temporal order of $k$.

\textbf{Dynamic Forest Model \citep{lehrmann2014efficient} (DFM).}  To compare our discriminative forest approach with a generative one, our third baseline is the a generative forest  where $k$ is the order of their non-linear Markov model. With no public implementation available, we directly report results in \citep{lehrmann2014efficient}.

\textbf{Pairwise Conditional Random Forest  \citep{dapogny2015pairwise} (PCRF).} To assess the discriminative pairwise information, we implement a pairwise forest similar to the one used for expression recognition \citep{dapogny2015pairwise}. We grow and combine classification trees for different pairwise temporal distance up to $k$. 

\subsection{Action recognition experiments}
We evaluate the proposed algorithm on three different action recognition benchmarks: MSRC-12 \citep{msrc12}, MSR-Action3D \citep{li2010action} and Florence-3D \citep{florence3d}. First, we perform detailed control experiments and parameter evaluation on MSRC-12 dataset. Next, we evaluate our approach comparing with baselines and state-of-the-art on all datasets.
\subsubsection{MSRC-12 experiments}
The MSRC-12 \citep{msrc12} dataset consists of 12 iconic and metaphoric gestures performed by 30 different actors. We follow the experimental protocol in \citep{lehrmann2014efficient}: only the 6 iconic gestures are used, making a total of 296 sequences and we perform 5-fold leave-person-out cross-validation, \ie, 24 actors for training and 6 actors for testing per fold.
\begin{figure}
\includegraphics[width=\columnwidth]{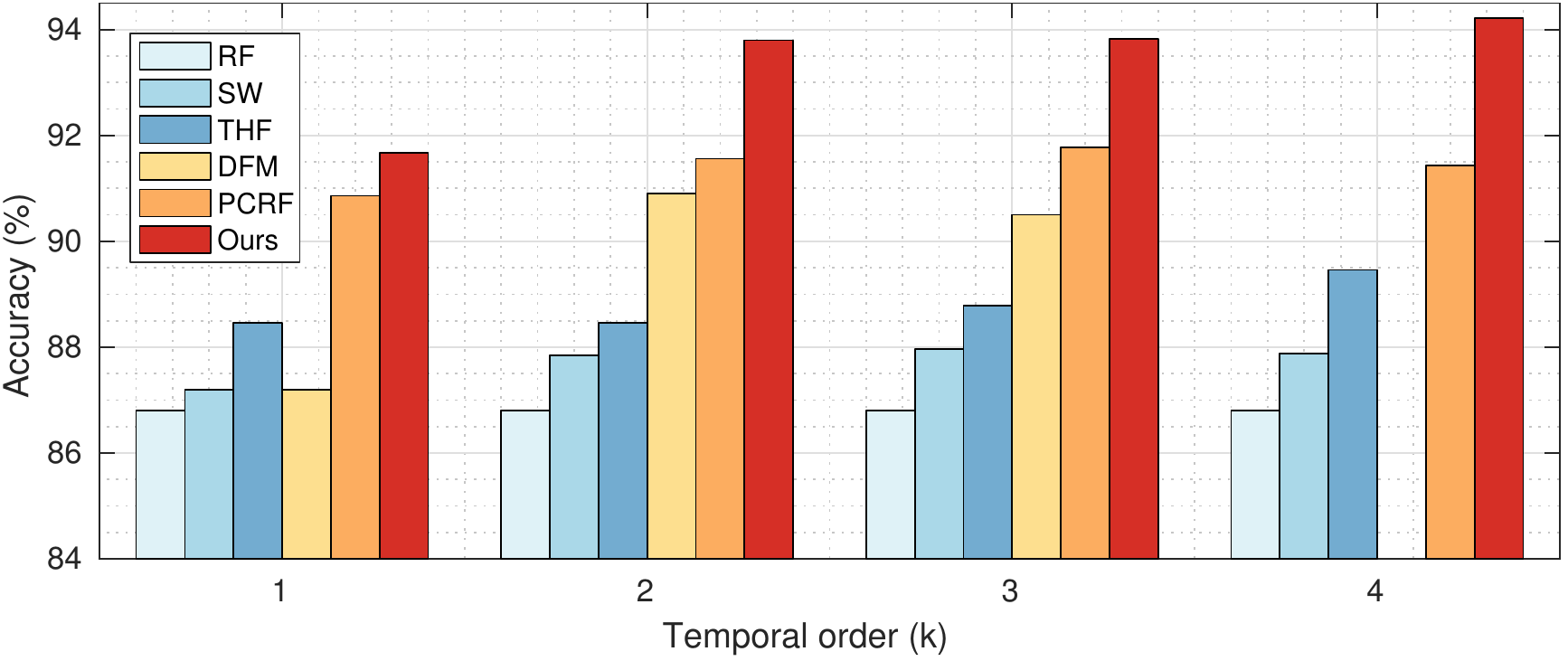}
\caption{Temporal order $k$ for different baselines and our approach on MSRC-12 dataset.}
\label{fig:msrc_baselines}	
\end{figure}
\begin{figure}
\includegraphics[width=\columnwidth]{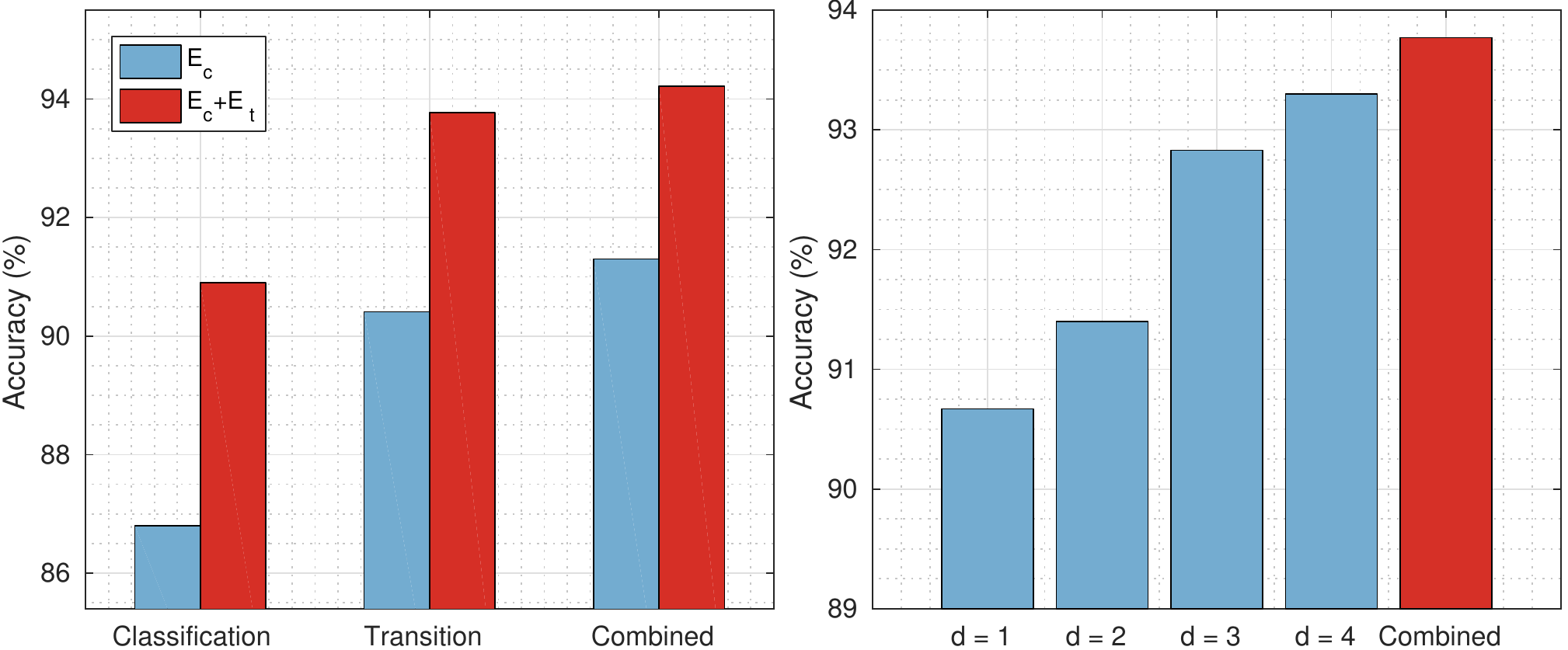}
\caption{(a) $E_c$ \vs $E_c+E_t$ and terms in Eq. \ref{eq:prediction}. (b) contribution of different $d$ order trees to transition probability shown in (a) and defined in Eq. \ref{eq:trans} on MSRC-12. }
\label{fig:msrc_transitions}	
\end{figure}

\textbf{Temporal order $k$ and comparison with baselines.} In Fig. \ref{fig:msrc_baselines} we show experimental results varying the temporal order parameter $k$ for all approaches. We observe that using only static information on single frames (RF) to recognize action is limited and it can be improved by stacking multiple frames (SW). Adding a regression term as in THF helps to increase the accuracy. DFM uses the same exact input window as SW, while being more robust as a result of their explicit modeling of time. Better than the rest of baselines, PCRF shows that capturing pairwise information is effective to model the temporal dynamics of the actions. On the other hand, our approach shows the best performance for all temporal orders. This shows that both combining static and temporal information in a discriminative way is very effective. In the next two paragraphs we analyze the contribution of both sources of information. 

\textbf{Discriminative power of learned transitions. } We measure the impact of our transition training procedure presented in Section \ref{Subsec: learning}. For this, we train two different transition forests, one using only $E_c$ and one using $E_c$ and $E_t$. For each forest, we show the performance by breaking down the terms of Eq. \ref{eq:prediction}: (i) using only the classification probability; (ii) using only the transition probability (Eq. \ref{eq:trans}); (iii) combining both terms (Eq. \ref{eq:prediction}). 

Results are shown in Fig. \ref{fig:msrc_transitions} (a). We observe that our proposed training algorithm increases the performance of both static and transition terms, leading to an important overall improvement. The static classification term improves substantially, meaning that $E_t$ helps to separate categories on the feature space by introducing temporal information that was not available otherwise. In Fig. \ref{fig:msrc_transitions} (b) we show the contribution of each temporal distance  to the overall transition probability in Eq. \ref{eq:trans}.
\begin{table}[t]
  \centering
  \resizebox{\columnwidth}{!}{%

  \begin{tabular}{lcccc}
      \toprule
  Method & Year & Real-time & Online  & Acc (\%) \\
  \midrule
  	DFM \cite{lehrmann2014efficient} & 2014 &  \checkmark & \checkmark & 90.90\\
  	ESM  \citep{jung2014enhanced} & 2014  & \xmark & \xmark & \textbf{96.76} \\
  	Riemann \cite{devanne2015combined} & 2015 & \xmark & \xmark & 91.50 \\
  	PCRF (our result) \cite{dapogny2015pairwise} & 2015 &  \checkmark & \checkmark & 91.77\\
  	Bag-of-poses \cite{zhu2016human} & 2016 &  \xmark & \xmark & 94.04 \\
    \midrule
    Ours (JP) & 2016 & \checkmark & \checkmark & \textbf{94.22}\\  
	Ours (RJP) & 2016 & \checkmark & \checkmark & \textbf{97.54}\\    
    Ours (MP) & 2016 & \checkmark & \checkmark & \textbf{98.25}\\    
  \bottomrule
  \end{tabular}
    }
	\caption{MSRC-12: Comparison with state-of-the-art using different frame representations.}
    \label{table:msrc-soa}
\end{table}

\textbf{Frame representation.} In addition to joint positions (JP) from above experiments, we experimented with two different frame representations: one static and one dynamic. The static one consists of pairwise relative distance of joints (RJP), proven to be more robust than JP while being very simple \citep{vemulapalli2014human}. The dynamic one, named Moving Pose (MP) \citep{Zanfir_2013_ICCV} incorporates temporal information by adding velocity and acceleration of joints using nearby frames. In Table \ref{table:msrc-soa} we observe that RJP and MP perform similarly well performing better than JP, showing that our approach can benefit of different static and dynamic feature representations.

\textbf{Initialization.} We initialized the transition nodes $N_t$ in two ways: randomly and using $E_c$. We found that the latter initialization provided slightly better results by $0.35\%$ after ten iterations. However, after doubling the number of iterations, the difference was reduced to $0.07\%$, leading to the conclusion that our algorithm is robust to initialization, but correctly initializing reduces the training time. Based on this, we limited the number of iterations to ten.

\textbf{Ensemble size.} A single tree of maximum depth 10 gave us an accuracy of $86.42\%$, six trees $93.10\%$ and twelve $94.22\%$. As a tree-based algorithm, adding more trees is expected to increase the performance (up to saturation) at the cost of computational time.

\textbf{Comparison with the state-of-the-art.} In Table \ref{table:msrc-soa} we compare our approach with the state-of-the-art. We observe that using the simple JP representation, we achieve the best with the exception of ESM \citep{jung2014enhanced}. However, ESM uses a slow variant of DTW and MP representation. Using both RJP and MP representation our approach achieves the best performance while being able to run in real time (1778 fps). 

\subsubsection{MSR-Action3D experiments.}

The MSR-Action3D \citep{li2010action} dataset is composed of 20 actions performed by 10 different actors. Each actor performed every action two or three times for a total of 557 sequences. We perform our main experiments following the setting proposed by \citep{li2010action}. In this protocol, the dataset is divided into three subsets of eight actions, named AS1, AS2 and AS3. The classification is performed on each subset separately and the final classification accuracy is the average over the three subsets. We perform a cross-subject validation in which half of the actors are used for training and the rest for testing using ten different splits. We use RJP frame representation, $k=4$ and 50 trees of maximum depth 8.

Baselines and state-of-the-art comparison are shown in Tables \ref{table:baselines} and \ref{table:action3dtable} respectively. Our approach achieves better performance than all baselines. Offline state-of-the-art methods \citep{zhangefficient,wang2016mining} achieve the best performance. Focusing on methods that are both real-time and online, the best performance is achieved by HURNN-L \citep{du2015hierarchical}, which uses a deep architecture to learn an end-to-end classifier. We obtain better results than \citep{du2015hierarchical} on both their online and offline flavors.

Some authors \cite{Zanfir_2013_ICCV,Veeriah_2015_ICCV} show results using a different protocol \citep{action3d} in which all 20 actions are considered. For comparison, using this protocol we achieved an accuracy of 92.8\%, which is superior to state-of-the-art online approaches of MP \citep{Zanfir_2013_ICCV}, 91.7\%, and dLSTM \citep{Veeriah_2015_ICCV}, 92.0\%, but inferior to the offline approach of Gram matrix \citep{zhangefficient}, 94.7\%. It is important to note that the inference complexity of both  \citep{Zanfir_2013_ICCV,zhangefficient} increases with the number of different actions, which is not the case of our approach, making it more suitable for realistic scenarios. \citep{zhangefficient} reported a testing time (ten runs over whole testing set) of 1523 seconds, for the same setting we report a significant lower time of 289 s.
\begin{table}[t]
  \centering
  \begin{tabular}{lccc}
      \toprule
  Method & MSRC-12 & MSR-Action3D & Florence-3D \\
  \midrule
    RF \cite{breiman2001random} & 86.83 &  87.77 & 85.46 \\
    SW \citep{msrc12} & 87.81 & 90.48 & 88.44\\	
    THF \citep{WACV} & 89.46 & 91.31 & 89.06 \\
	DFM \citep{lehrmann2014efficient} & 90.90 & - & - \\ 
	PCRF \citep{dapogny2015pairwise} & 91.77 & 92.09 & 91.23 \\ 
    \midrule
    Ours & \textbf{94.22}& \textbf{94.57} & \textbf{94.16} \\   
  \bottomrule
  \end{tabular}
    \caption{Comparison with forest-based baselines.}
      \label{table:baselines}
\end{table}

\begin{table}
  \centering
  \resizebox{\columnwidth}{!}{%

  \begin{tabular}{lcccc}
      \toprule
  Method & Year & Real-time & Online  & Acc (\%) \\
  \midrule
  	Bag of poses \cite{florence3d} & 2013 &  \xmark & \xmark & 82.15 \\
    Lie group \citep{vemulapalli2014human} & 2014 & \xmark & \xmark & 90.88 \\
    PCRF (our result) \cite{dapogny2015pairwise} & 2015 &  \checkmark & \checkmark & 91.23\\
    Rolling rot. \citep{rolling} & 2016 & \xmark & \xmark & 91.40 \\
	Graph-based \citep{wang2016graph} & 2016 & \xmark & \xmark & 91.63 \\
	Key-poses \citep{wang2016mining} & 2016 & \checkmark & \xmark & 92.25 \\

    \midrule
    Ours & 2016 & \checkmark & \checkmark & \textbf{94.16}\\    
  \bottomrule
  \end{tabular}
    }
        \caption{Florence-3D: Comparison with state-of-the-art.}
          \label{table:f3-table}
\end{table}

\subsubsection{Florence-3D experiments}

The Florence-3D dataset \cite{florence3d} consists of 9 different actions performed by 10 subjects. Each subject performed every action two or three times making a total of 215 action sequences. Following previous work \citep{wang2016mining,wang2016graph}, we adopt a leave-one-subject-out protocol, \eg nine subjects are used for training and one for testing for ten times. We used the same parameters as in the previous experiment.

We compare the proposed approach with baselines and state-of-the-art in Tables \ref{table:baselines} and \ref{table:f3-table} respectively. We can see that our approach achieves the best performance over all baselines and state-of-the-art. Note that on this dataset we outperform the recent Key-poses approach \citep{wang2016mining}, which achieved the best performance on MSR-Action3D dataset. 

\begin{table*}[t]
  \centering
  \begin{tabular}{lccccccc}
      \toprule
  Method & Year & Real-time & Online & AS1 (\%) & AS2 (\%) & AS3 (\%) & Average (\%) \\
  \midrule
  	BoF forest \citep{zhu2013fusing} & 2013 & \xmark & \xmark & - & - & - & 90.90 \\
    Lie group \citep{vemulapalli2014human} & 2014 & \xmark & \xmark & 95.29 & 83.87 & 98.22 & 92.46 \\
	HBRNN-L \citep{du2015hierarchical} & 2015 & \checkmark & \xmark & 93.33 & 94.64 & 95.50 & 94.49 \\ 
	Graph-based \citep{wang2016graph} & 2016 & \xmark  & \xmark & 93.75 & 95.45 & 95.10 & 94.77 \\
	Gram matrix \citep{zhangefficient} & 2016 & \checkmark & \xmark & 98.66 & 94.11 & 98.13 & 96.97 \\
	Key-poses \cite{wang2016mining} & 2016 & \checkmark & \xmark & - & - & - & \textbf{97.44} \\

	\midrule
	PCRF (our result) \cite{dapogny2015pairwise} & 2015 &  \checkmark & \checkmark & 94.51 & 85.58 & 96.18 & 92.09\\
	HURNN-L \citep{du2015hierarchical} & 2015 & \checkmark &  \checkmark & 92.38 & 93.75 & 94.59 & \textbf{93.57} \\ 
    \midrule

    Ours & 2016 & \checkmark & \checkmark & 96.10 & 90.54 & 97.06 & \textbf{94.57}\\    
  \bottomrule
  \end{tabular}
    \caption{MSR-Action3D: Comparison with state-of-the-art.}
      \label{table:action3dtable}

\end{table*}

\subsection{Online action detection experiments}
\begin{table*}[t]
  \centering
  \begin{tabular}{lccccccc}
\toprule  
&
\multicolumn{3}{c}{Baselines}    &
\multicolumn{2}{c}{State-of-the-art}   &
&   \\ 
\cmidrule(lr){2-4}
\cmidrule(lr){5-6}

Action &
\multicolumn{1}{c}{RF} &
\multicolumn{1}{c}{SW}     &
\multicolumn{1}{c}{PCRF} &
\multicolumn{1}{c}{RNN \cite{zhu2016co}}    &
\multicolumn{1}{c}{JCR-RNN \cite{li2016online} }     &
Ours\\
 \midrule
 drinking 						& 0.598 & 0.387 & 0.468 &  0.441 &  0.574 & \textbf{0.705} \\
 eating 							& 0.683	& 0.590 & 0.550 &  0.550 &  0.523 & \textbf{0.700} \\
 writing 						& 0.640 & 0.678 & 0.703 &  \textbf{0.859} &  0.822 & 0.758 \\
 opening cupboard 				& 0.367	& 0.317 & 0.303 &  0.321 &  \textbf{0.495} & 0.473 \\
 washing hands 					& 0.698	&\textbf{0.792} & 0.613 &  0.668 &  0.718 & \textbf{0.740} \\
 opening microwave 				& 0.525	& 0.717 & 0.717 &  0.665 &  0.703 & \textbf{0.717} \\
 sweeping 						& 0.539	& 0.583 & 0.635 &  0.590 &  0.643 & \textbf{0.645} \\
 gargling  						& 0.298	& 0.414 & 0.464 &  0.550 &  0.623 & \textbf{0.633} \\
 throwing trash 					& 0.340 & 0.205 & 0.350 & \textbf{0.674} &  0.459 & 0.518 \\
 wiping 							& 0.823 & 0.765 & 0.823 &  0.747 &  0.780 & \textbf{0.823} \\
 \midrule
 Overall 						& 0.578	& 0.556 & 0.607 & 0.600  & 0.653 & \textbf{0.712}\\
  \midrule
SL 								& 0.361	& 0.366 & 0.378 & 0.366 & 0.418 & \textbf{0.514} \\
EL 								& 0.391	& 0.326 & 0.412 & 0.376 & 0.443 & \textbf{0.527} \\
\midrule

Inference time (s)			   & 0.59 &  0.61  &  3.58  &  3.14 & 2.60 & \textbf{1.84} \\
  \bottomrule

  \end{tabular}
    \caption{Performance comparison on Online Action Detection (OAD) dataset.}
  \label{table:oad}
\end{table*}

We end our experimental evaluation in a more realistic scenario. We test our approach for online action detection on the very recently proposed Online Action Detection (OAD) dataset \cite{li2016online}. The dataset consists of 59 long sequences containing 10 different daily-life actions performed by different actors. Each sequence contains different action/background periods of variable length in arbitrary order annotated with start/end frames. We use the same splits and evaluation protocol as \cite{li2016online}. 
Previous work \cite{li2016online} fixed the number of considered previous frames to $10$, in consequence we set $k=10$. We use RJP representation and 50 trees of maximum depth 20. Thresholds $\beta_s$ and $\beta_e$ were empirically set to $0.79$ and $0.16$ respectively.

In Table \ref{table:oad} we report class-wise and overall F1-score for baselines, state-of-the-art and our approach. We also report the accuracy of start and end frame detection `SL' and `EL' respectively. We observe that our approach outperforms all baselines. PCRF forest shown the best results among the baselines with a performance comparable to RNN, showing that temporal pairwise information is important. On the other hand, RF performs particularly well on this dataset, revealing that distinguishing static poses is important in addition to temporal information. Combining both static and temporal information in our approach led us to better performance than the current state-of-the-art JCR-RNN \cite{li2016online}, which added a regression term on a LSTM to predict both start and end frames of actions.
 
\textbf{Efficiency}. We measure the average inference time on 9 long sequences of 3200 frames in average. We present the results at the bottom of Table \ref{table:oad} with a C++ implementation on a Intel Core i7 (2.6 GHz) and 16 GB RAM. All compared approaches are real-time, with JCR-RNN achieving 1230 fps for 1778 fps of our approach, showing that we can obtain high performance while keeping the complexity low. 
\section{Summary and conclusion}
\label{Sec: conclusion}
We proposed a new forest based classifier that is able to learn both static poses and transitions in a discriminative way. Our proposed training procedure helps to capture temporal dynamics in a more effective way than other strong forest baselines. Introducing temporal relationships while growing the trees and also using them in inference helped to obtain more robust frame-wise predictions, leading us to show state-of-the-art performance in both challenging problems of action recognition and online action detection.  

Currently, our learning stage is limited to pairwise transitions and we believe that it would be interesting to incorporate different time orders within the same tree learning. Also, given the generality of our work, it would be interesting to test its performance using other data modalities (such as RGB/depth frame features) or applied to other temporal problems requiring efficient and online classification.

{\small
\bibliographystyle{ieee}
\bibliography{main.bib}
}

\end{document}